\documentclass[11pt]{article}
\usepackage{coling2020}
\usepackage{times}
\usepackage{url}
\usepackage{latexsym}

\usepackage{graphicx}
\usepackage{xspace}
\usepackage[hidelinks]{hyperref}
\usepackage{booktabs}
\usepackage{microtype}
\usepackage{tabularx}
\usepackage{enumitem}
\usepackage{amssymb}
\usepackage{wrapfig}
\usepackage{caption}
\usepackage{multirow}
\usepackage{xspace}
\newcommand{\emotionname}[1]{\textit{#1}}
\newcommand{\fear}{\emotionname{fear}\xspace}
\newcommand{\joy}{\emotionname{joy}\xspace}
\newcommand{\anger}{\emotionname{anger}\xspace}
\newcommand{\trust}{\emotionname{trust}\xspace}
\newcommand{\guilt}{\emotionname{guilt}\xspace}
\newcommand{\shame}{\emotionname{shame}\xspace}
\newcommand{\surprise}{\emotionname{surprise}\xspace}
\newcommand{\sadness}{\emotionname{sadness}\xspace}
\newcommand{\anticipation}{\emotionname{anticipation}\xspace}
\newcommand{\disgust}{\emotionname{disgust}\xspace}
\newcommand{\pleasantness}{\emotionname{pleasantness}\xspace}
\newcommand{\effort}{\emotionname{anticipated effort}\xspace}
\newcommand{\certainty}{\emotionname{certainty}\xspace}
\newcommand{\attention}{\emotionname{attention}\xspace}
\newcommand{\responsibilityC}{\emotionname{self-other responsibility/control}\xspace}
\newcommand{\responsibility}{\emotionname{responsibility}\xspace}
\newcommand{\situationalControl}{\emotionname{situational control}\xspace}
\newcommand{\control}{\emotionname{control}\xspace}
\newcommand{\circumstance}{\emotionname{circumstance}\xspace}
\newcommand{\avg}{avg.}
\newcommand{\F}{$\textrm{F}_1$\xspace}
\newcommand{\rt}[1]{\rotatebox{90}{#1}}
\newcommand{\taskTE}{T$\rightarrow$E\xspace}
\newcommand{\taskTA}{T$\rightarrow$A\xspace}
\newcommand{\taskAE}{A$\rightarrow$E\xspace}
\newcommand{\taskTAE}{T$\rightarrow$A/E\xspace}

\usepackage{trimclip}
\newcommand{\shortarrow}{\clipbox{5pt 0pt 0pt -1pt}{$\rightarrow$}}

\colingfinalcopy

\title{Appraisal Theories for Emotion Classification in Text}

\author{Jan Hofmann$^1$, Enrica Troiano$^1$, Kai Sassenberg$^{2,3}$, \and Roman Klinger$^1$ \\
  $^1$Institut f{\"u}r Maschinelle Sprachverarbeitung, University of Stuttgart, Germany \\
  $^2$Leibniz-Institut f\"ur Wissensmedien, T\"ubingen, Germany\\
  $^3$University of T\"ubingen, Germany\\
  \texttt{\{jan.hofmann,enrica.troiano,roman.klinger\}@ims.uni-stuttgart.de}\\
  \texttt{k.sassenberg@iwm-tuebingen.de}\\
}

\date{}

\begin{document}
\maketitle
\begin{abstract}
  Automatic emotion categorization has been predominantly formulated
  as text classification in which textual units are assigned to an
  emotion from a predefined inventory, for instance following the
  fundamental emotion classes proposed by Paul Ekman (\fear, \joy,
  \anger, \disgust, \sadness, \surprise) or Robert Plutchik (adding
  \trust, \anticipation). This approach ignores existing psychological
  theories to some degree, which provide explanations regarding the
  perception of events. For instance, the description that somebody
  discovers a snake is associated with \fear, based on the appraisal
  as being an unpleasant and non-controllable situation. This emotion
  reconstruction is even possible without having access to explicit
  reports of a subjective feeling (for instance expressing this with
  the words ``I am afraid.''). Automatic classification approaches
  therefore need to learn properties of events as latent variables
  (for instance that the uncertainty and the mental or physical effort associated with
  the encounter of a snake leads to \fear). With this paper, we propose
  to make such interpretations of events explicit, following theories
  of cognitive appraisal of events, and show their potential for
  emotion classification when being encoded in classification
  models. Our results show that high quality appraisal dimension
  assignments in event descriptions lead to an improvement in the
  classification of discrete emotion categories. We make our corpus of
  appraisal-annotated emotion-associated event descriptions publicly
  available.
\end{abstract}

\blfootnote{
  \hspace{-0.65cm}  
  This work is licensed under a Creative Commons 
  Attribution 4.0 International License.
  License details:
  \url{http://creativecommons.org/licenses/by/4.0/}.
}

\section{Introduction}
The task of emotion analysis is commonly formulated as classification
or regression in which textual units (documents, paragraphs,
sentences, words) are mapped to a predefined reference system, for
instance the sets of fundamental emotions \textit{fear},
\textit{anger}, \textit{joy}, \textit{surprise}, \textit{disgust}, and
\textit{sadness} proposed by \newcite{Ekman1999}, or by
\newcite{Plutchik2001}, which includes also \textit{trust} and
\textit{anticipation}. Machine learning-based models need to figure
out which words point to a particular emotion experienced by
a reader, by the author of a text, or a character in it. Depending on the
resource which has been annotated, the description of an emotion
experience can vary.  On Twitter, for instance, other than direct
reports of an emotion state (``I feel depressed''), hashtags are used
as emotion labels to enrich the description of events and stances (``I
just got my exam result \#sad''). In news articles, emotional events
are sometimes explicitly mentioned (``couple infuriate officials''
\cite{Bostan2020}) and other times require world knowledge (``Tom
Cruise and Katie Holmes set wedding date'', labeled as \surprise
\cite{Strapparava2007}). In literature, a sequence of events which
forms the narrative leads to an emotion in the reader. In this paper,
we focus on those texts which communicate emotions without an explicit
emotion word, but rather describe events for which an emotion
association is evident.

Such textual examples became popular in natural language processing
research with the use of the data generated in the
ISEAR project \cite{Scherer1997}. The project led to a dataset of
descriptions of events triggering specific affective states, which was
originally collected to study event interpretations with a psychological
focus. In text analysis, to infer the emotion felt by the writers of
those reports, an event interpretation needs to be accomplished. For
instance, in the text ``When a car is overtaking another and I am
forced to drive off the road'', the model needs to associate the event
with \fear.

To date, nearly all computational approaches that associate text with
emotions are agnostic to the way in which emotions are communicated, they
do ``not know'' how to interpret events, but, presumably, they purely
learn word associations instead of actual event interpretations. One
might argue that approaches predicting fine-grained
dimensions of affect, namely arousal and valence, actually tackle this problem
\cite{Buechel2017,Preotiuc2016}. However, these typically do not infer
downstream emotion categories. Further, particularly regarding events,
psychological theories offer more detailed information. As an example,
the emotion component model \cite{Scherer2005} advocates that
cognitive appraisal dimensions underly discrete emotion classes
\cite{Smith1985}. These appraisal dimensions\footnote{The examples
  follow the results by \newcite{Smith1985}, an excerpt is shown in
  Table~\ref{tab:smithelssworth}.} evaluate (1) how pleasant an event
is (\pleasantness, likely to be associated with \joy, but unlikely to
appear with \disgust), (2) how much effort an event can be expected to
cause (\effort, likely to be high when \anger or \fear is
experienced), (3) how certain the experiencer is in a specific
situation (\certainty, low, e.g., in the context of \emotionname{hope}
or \emotionname{surprise}), (4) how much attention is devoted to the
event (\attention, likely to be low, e.g., in the case of
\emotionname{boredom} or \disgust), (5) how much
responsibility the experiencer of the emotion holds for what has
happened (\responsibilityC, high for feeling \emotionname{challenged}
or \emotionname{pride}), and (6) how much the experiencer has control
over the situation (\situationalControl, low in the case of \anger).

Despite their richness, cognitive theories of appraisal and their
empirical results have not been exploited for emotion prediction in
text yet. We fill this gap with this paper and analyze the relation
between appraisal dimensions and emotion categories in a text
classification setting. We post-annotate an English emotion corpus of
self-reports of emotion events \cite{Troiano2019}, which already
contains annotations related to the emotions of \anger, \disgust,
\fear, \guilt, \joy, \sadness, and \shame, and add the appraisal
dimensions by \newcite{Smith1985} mentioned above. Further, we analyze
if an automatic prediction of these dimensions from text is possible
with standard neural methods, and if these predictions contribute to
emotion classifications. Our main contributions are: (1) the first
event-centered text corpus annotated with appraisal dimensions; (2)
the evaluation of how well text classification models can recognize these
appraisal dimensions; (3) we show emotion classification benefits from
the information of appraisal dimensions, when high quality predictions
of these are available. Further, (4), we replicate the study by
\newcite{Smith1985} from a CL/NLP perspective,
based on textual event descriptions.

\section{Background on Emotion Psychology and Analysis}
\label{sec:relatedwork}
\subsection{Emotion and Affect Theories}
As a component of humans' life, emotions have been thoroughly
studied in the field of psychology, where they are generally deemed
responses to salient events. The debates surrounding their definition,
however, has never come to a consensus, producing a varied literature
on the topic. This has a clear implication for computational emotion
analyses, for they must choose and follow one of the available
psychological theories in order to motivate the emotion phenomenon
that they research in language.

Some of such theories focus on the evolutionary function of emotions,
and accordingly, on their link to actions \cite{izard1971,tooby2008}.
The core idea is that emotions help humans accomplish every-day life
tasks and communicate socially relevant information by triggering
specific physiological symptoms. In particular, there are patterns of
behaviour (e.g., smiling) that reflect discrete emotion terms (e.g.,
\joy), which suggests that emotional states can be grouped based on a
few natural language categories.  One of the most popular sources for
a set of fundamental emotions is the theory by \newcite{Ekman1992}.
Ekman studied the relation between emotions and both culture and
facial expressions: he claimed that the set of fundamental emotions,
namely, \anger, \disgust, \fear, \joy, \sadness, and \surprise can be
distinguished by facial muscular movements across cultures (which is
partially doubted these days, \newcite{Gendron2014}). As an addition
to this model, \newcite{Plutchik2001} makes explicit the assumption
that different fundamental emotions can occur together, for instance
\trust and \joy, which is the case when \emotionname{love} is
experienced. Such emotion mixtures, as well as an opposition between
\anger and \fear, \joy and \sadness, \surprise and \anticipation,
\trust and \disgust, has been included in this model.  In natural
language processing, mostly a set of four to eight fundamental
emotions is used, where \anger, \fear, \joy, and \sadness are shared
by most approaches (an exception with 24 emotion classes is
\newcite{Ungar2017}).

A diametrically opposite view is held by the constructivist tradition
\cite{averill1980,oatley1993,barrett2015}, in which actions and
physiological changes are the building blocks that construct emotions,
rather than their direct effect \cite{Barrett2006}. Feeling an emotion
means categorizing the fluctuations of an affect system along some
components.  For instance, the affect components \emph{valence}
(degree of polarity), \emph{arousal} (degree of excitement), and
\emph{dominance} (degree of control over a situation)
\cite{Posner2005} are used as dimensions to describe
affect experiences in a 3-dimensional space, which can then be mapped
to discrete emotion categories.

\subsection{Theories of Cognitive Appraisal}
Extensions to this model in terms of underlying components are the
works of \newcite{scherer1982}, \newcite{Smith1985} and
\newcite{oatley1987}, who qualified emotions as \textit{component}
processes that arise to face salient circumstances: an emotion is an
``episode of interrelated, synchronized changes in the states of all
or most of the five organismic subsystems in response to the
evaluation of an external or internal stimulus-event as relevant to
major concerns of the organism'' \cite{scherer2001}.  According to
this view, there is an appraisal, that is, an information processing
component, which enables people to determine the significance of a
situation with respect to their needs and values. In the context of
this appraisal (e.g., judging a snake as dangerous), the resources of
four other components are mobilitated to deal with the situation.
These are, next to the \emph{cognitive component (appraisal)}, a
\emph{neurophysiological component (bodily symptoms)}, a
\emph{motivational component (action tendencies)}, a \emph{motor
  expression (facial and vocal expression)}, and a \emph{subjective
  feeling component (emotional experience)}
\cite[Table~1]{Scherer2005}.

While the notions of subjective experience and bodily symptoms were
common to other emotion theories, appraisal represents a
novelty that fills in some shortcomings of basic models.
First, it explains how emotions are elicited. The origin of
emotions is to be seen in the stimulus as appraised rather than in the
stimulus as such. Second, appraisals provide a structured account for
the differences among emotions. For instance, \anger and \fear are
experienced when the evaluation of a negative event
attributes it to external factors, whereas \guilt and \shame are felt
if the causes of such event are identified in the self, as stable and
uncontrollable personality traits, like in the case of \shame
(e.g., ``I'm dumb''), or unstable and controllable behaviours for
\guilt (e.g., ``I did not observe the speed limit'') \cite{tracy2006}.

We argue in this paper that this makes appraisals particularly useful
for natural language processing, because they both provide a framework
for research and represent a way of enriching existing data. As a
matter of fact, few dimensions are sufficient to explain emotions
based on cognitive appraisal.  \newcite{Smith1985} explain 15 emotions
by leveraging \pleasantness (polarity), \responsibilityC (for initiating
the situation), \certainty (about what is going on), \attention
(whether the emotion stimulus is worth attending), \effort (the amount
of physical or mental activation before the stimulus), and
\situationalControl (the ability to cope with the situation). Compared
to the valence-arousal-dominance model, where it is left unclear if
the polarity dimension refers to a quality of the emotion stimulus or
a quality of the feeling \cite{Scherer2005}, all these dimensions are
unambiguously event-directed.
In this paper, we focus on modelling the cognitive components
described by \newcite{Smith1985}. We show their main findings in
Table~\ref{tab:smithelssworth}, limited to the emotions that are
available in the corpus we use.

\begin{table}
  \centering\footnotesize
  \begin{tabular}{lrrrrrr}
    \toprule
    Emotion & Unpleasant & Responsibility & Uncertainty & Attention & Effort & Control \\
    \cmidrule(r){1-1}\cmidrule(rl){2-2}\cmidrule(rl){3-3}\cmidrule(rl){4-4}\cmidrule(rl){5-5}\cmidrule(rl){6-6}\cmidrule(l){7-7}
    Happiness & $-$1.46 & 0.09 & $-$0.46 & 0.15 & $-$0.33 & $-$0.21 \\
    Sadness   & 0.87 & $-$0.36 & 0.00 & $-$0.21 & $-$0.14 & 1.15\\
    Anger     & 0.85 & $-$0.94 & $-$0.29 & 0.12 & 0.53 & $-$0.96  \\
    Fear      & 0.44 & $-$0.17 & 0.73 & 0.03 & 0.63 & 0.59\\
    Disgust   & 0.38 &$-$0.50 &$-$0.39 &$-$0.96 &0.06 &$-$0.19\\
    Shame     & 0.73 &1.31 &0.21 &$-$0.11 &0.07 &$-$0.07\\
    Guilt     & 0.60 &1.31 &$-$0.15 &$-$0.36 &0.00 &$-$0.29 \\
    \bottomrule
  \end{tabular}
  \caption{The locations of emotions along appraisal dimensions
    (according to a principle component analysis) as
    published by \protect\newcite{Smith1985}, Table 6, filtered to
    those emotions which are available in the text corpus we use.}
  \label{tab:smithelssworth}
\end{table}

\subsection{Automatic Emotion Classification}
Previous work on emotion analysis in natural language processing
focuses either on resource creation or on emotion classification for a
specific task and domain. On the side of resource creation, the early
and influential work of \newcite{Pennebaker2001} is a dictionary of
words being associated with different psychologically relevant
categories, including a subset of emotions. Later,
\newcite{Strapparava2004} made WordNet Affect available to target word
classes and differences regarding their emotional connotation,
\newcite{Mohammad2012b} released the NRC dictionary with more than
14,000 words for a set of discrete emotion classes, and a
valence-arousal-dominance dictionary was provided by
\newcite{Mohammad2018vad}. \newcite{Buechel2016} have developed a
methodological framework to adapt existing affect lexicons to specific
use cases.  Other than dictionaries, emotion analysis relies on
labeled corpora.  Some of them include information relative to valence
and arousal \cite{Buechel2017,Preotiuc2016}, but the majority of
resources use discrete emotion classes, for instance to label fairy
tales \cite{Ovesdotter2005}, blogs \cite{Aman2007}, tweets
\cite{Mohammad2017c,Schuff2017,Mohammad2012,Mohammad2017,Klinger2018},
Facebook posts \cite{Preotiuc2016}, news headlines
\cite{Strapparava2007}, dialogues \cite{Li2017}, literary texts
\cite{Kim2017a}, or self reports on emotion events
\cite{Scherer1997,Troiano2019}. We point the reader to the survey by
\newcite{Bostan2018} for a more an overview.

Most automatic methods that assign labels to text rely on machine
learning \cite[i.a.]{Ovesdotter2005,Aman2007,Schuff2017}. Recent
shared tasks showed an increase in transfer learning from generic
representations
\cite{Klinger2018,Mohammad2018,Mohammad2017}. \newcite{Felbo2017}
proposed to use emoji representations for pretraining, and
\newcite{Cevher2019} performed pretraining on existing emotion corpora
followed by fine-tuning for a specific domain for which only little
training data was available.

We are only aware of one preliminary study which considered appraisal
dimensions to improve text-based emotion prediction, namely
\newcite{Campero2017}. In their study, subjects labeled 200 stories
with 38 appraisal features (which remain unmentioned), to evaluate if
a text-based representation adds on top of an fMRI-based
classification. Apart from this study, all previous machine
learning-based approaches used models to predict emotions or affect
values directly from text, without any access to appraisal dimensions.
Only a couple of works incorporated cognitive components, for instance
those coming from the OCC model (named after the authors Ortony, Clore
and Collins' initials), which sees every appraisal as an evaluation of
the pleasantness of events, objects, or actions with respect to one's
goals, tastes or behavioural and moral standards \cite{Clore2013}.
Based on the OCC model, \newcite{Shaikh2009} devised a rule-based
approach to interpret text.  They did not explicitly formulate their
model following appraisal theories, but they moved towards a
cognitively-motivated interpretation of events and interpersonal
descriptions.
Others have adopted patterns of appraisal to predict the emotions
triggered by actions, as described in a text. Specifically,
\newcite{Balahur2011} and \newcite{Balahur2012} have created EmotiNet,
a knowledge base of action chains that includes information about the
elements on which the appraisal is performed within an affective
situation, namely, the agent, the action and the object involved in a
chain. We share their motivation to delve into event representations
based on the descriptions of their experiencers.
Unlike their work, ours explicitly encodes appraisal dimensions and uses
the classification into these categories for emotion prediction.

We are not aware of any textual corpus annotated with appraisals
according to emotion theories, or associated computational
models. However, some research exists which studies these theories
with added value for computational models. \newcite{Scherer2019b},
e.g., explicitly encode realizations of appraisals of events and
compare those between production and perception of facial
expressions. \newcite{Marsella2009} model emotions in video sequences
computationally, based on appraisals of events.
\newcite{Broekens2008} inform intelligent agents about emotions based
on formalizations of cognitive appraisal. Further, in
\newcite{Scherer2010}, the chapter authored by \newcite{Marsella2010} argues
for an integrated computational approach of emotions which also embeds
appraisal theories.

\section{Corpus}
\label{sec:corpus}
The main objective of this study is to understand the relation between
appraisal dimensions and emotion categories. Therefore, we build
appraisal annotations on top of enISEAR, an existing corpus of 1001
English event descriptions which are already labeled with the discrete
categories of \anger, \disgust, \fear, \guilt, \joy, \sadness, and
\shame \cite{Troiano2019}. Each instance has been generated by a
crowdworker on the platform FigureEight by completing the sentence ``I
feel [emotion name], when \ldots''. This corpus has an advantage over
the original ISEAR resource because it has a German counterpart which
can be used in further studies; moreover, its emotion labels have been
intersubjectively validated. The corpus, annotations, and our
implementations are available at
\url{http://www.ims.uni-stuttgart.de/data/appraisalemotion}.

\subsection{Annotation}
One presumable challenge in the post-annotation of events regarding
the appraisal dimensions is that our annotators do not
have access to the private state of the event experiencers. However, under the assumption that events are perceived
similarly in subjective feeling and evaluated comparably based on
cognitive appraisal, we assume that this is not a major flaw in the
design of the study. An alternative would have been to perform the
text generation task as \newcite{Troiano2019} did, but asking the
authors of event descriptions for their appraisal in addition. We
opted against such procedure as it would have meant to reproduce an
existing study in addition to our research goal.

For the post-labeling of enISEAR, we aimed at formulating unambiguous
and intuitive descriptions of appraisal dimensions, which would be
faithful to those in \newcite{Smith1985}. As opposed to the subjects
of their study, however, our annotators had to judge events that they
did not personally experience. For this reason, we simplified our
annotation guidelines in two respects. First, we opted for a binary
setting, while \newcite{Smith1985} used continuous scales to rate
discrete emotion categories on the appraisal dimensions. Second, we
split \control into \emph{Control} and \emph{Circumstance} (i.e.,
\emph{self} and \situationalControl), in line with the discussion of
this variable by \newcite[p.\ 824f.]{Smith1985}, while retaining the
category of \responsibility. This was motivated by the observation of
a low inter-annotator agreement in preliminary annotation rounds and a
series of discussions that revealed the difficulty for annotators to
separate the concepts of responsibility and self control. Then, the
annotators were instructed to read an event description, without
having access to
the emotion label, and to answer the following questions:\\[-3mm]

Most probably, at the time when the event happened, the writer\ldots
\begin{itemize}[topsep=0pt,itemsep=0pt,parsep=0pt,partopsep=0pt]
\item \ldots wanted to devote further attention to the event. (\textit{Attention})
\item \ldots was certain about what was happening. (\textit{Certainty})
\item \ldots had to expend mental or physical effort to deal with the
  situation. (\textit{Effort})
\item \ldots found that the event was pleasant. (\textit{Pleasantness})
\item \ldots was responsible for the situation. (\textit{Responsibility})
\item \ldots found that he/she was in control of the situation. (\textit{Control})
\item \ldots found that the event could not have been changed or
  influenced by anyone. (\textit{Circumstance})
\end{itemize}
Each event description from enISEAR was judged by
three annotators between the age of 26 and 29. One of them is a
female Ph.D.\ student of computational linguistics, the others are
male graduate students of software engineering. Two of the annotators
are co-authors of this paper. The judges familiarised themselves
with their task through four training iterations. At every iteration,
we hand-picked 15--20 samples from the ISEAR dataset
\cite{Scherer1997}, such that instances used for training would not be seen
during the actual annotation, but had a comparable structure.
Dissimilarities in the annotation were discussed in
face-to-face meetings and the annotation guideline was refined.

The agreement improved from $\kappa$=$0.62$ to 0.67 in the four
iterations. In one of them, we experimented with giving access to the
emotion label, which lead to a large improvement in agreement
($\kappa$=$0.83$). Nevertheless, we decided to continue
without this information, in order to evaluate the annotator's
performance in a similar setting as we evaluate the automatic model --
to predict appraisal for emotion classification.
We show the pairwise inter-annotator scores of the final set in
Table~\ref{tab:iaa}. The agreement scores between the different
annotator pairs are comparable.

\begin{table}
  \centering\small
  \setlength{\tabcolsep}{10pt}
  \begin{tabular}{lcccccccc}
    \toprule
    &\multicolumn{8}{c}{Cohen's $\kappa$}\\
    \cmidrule(l){2-9}
    & \multicolumn{4}{c}{between annotators} & \multicolumn{4}{c}{annotator--majority} \\
    \cmidrule(r){2-5}\cmidrule(l){6-9}
    Appraisal Dimension& A1/A2 & A1/A3 & A2/A3 & \avg & A1 & A2 & A3 & \avg\\
    \cmidrule(r){1-1}\cmidrule(lr){2-5}\cmidrule(l){6-9}
    Attentional Activity & .28 & .24 & .41 &.31 & .50 & .76 & .66 &.64\\
    Certainty            & .41 & .23 & .29 &.31 & .62 & .77 & .46 &.62 \\
    Anticipated Effort   & .38 & .33 & .26 &.32 & .69 & .67 & .62 &.66 \\
    Pleasantness         & .89 & .88 & .90 &.89 & .93 & .96 & .94 &.94 \\
    Responsibility       & .68 & .57 & .63 &.63 & .80 & .88 & .76 &.81 \\
    Control              & .65 & .56 & .52 &.58 & .84 & .81 & .70 &.78 \\
    Circumstance         & .52 & .32 & .28 &.37 & .80 & .69 & .49 &.66 \\
    \cmidrule(r){1-1}\cmidrule(lr){2-5}\cmidrule(l){6-9}
    Average              & .59 & .48 & .52 &.53 & .77 & .82 & .70 & .76\\
    \toprule
  \end{tabular}
  \caption{Cohen's $\kappa$ between all annotator pairs and between
    each annotator and the majority vote.}
  \label{tab:iaa}
\end{table}

These scores tell us that rating appraisal dimensions for given events
is challenging, and its difficulty varies depending on the categories.
Given the comparably low agreement obtained for a subset of
dimensions, we opt for a ``crowd-sourcing''-like aggregation by taking
the majority vote to form the final annotation, included in Table~\ref{tab:iaa},
on the right side of the table. We
observe that the agreement between majority vote and each annotator is
constantly above $\kappa$=.62, which is an acceptable agreement (\avg
$\kappa$=.76).

\newcommand{\mco}[1]{\multicolumn{2}{c}{#1}}
\begin{table}
  \centering\small
  \setlength{\tabcolsep}{6pt}
  \renewcommand{\arraystretch}{0.99}
    \begin{tabular}{rrrrrrrrrrrrrrr}
      \toprule
      & \multicolumn{14}{c}{Appraisal Dimension}\\
      \cmidrule(l){2-15}
      Emotion  & \mco{Attention} & \mco{Certainty} & \mco{Effort} & \mco{Pleasant} & \mco{Respons.} & \mco{Control} & \mco{Circum.}\\
      \cmidrule(r){1-1}\cmidrule(lr){2-3}\cmidrule(rl){4-5}\cmidrule(rl){6-7}\cmidrule(rl){8-9}\cmidrule(rl){10-11}\cmidrule(rl){12-13}\cmidrule(l){14-15}
Anger & 129 & .90 & 119 & .83 & 60 & .42 & 0 & .00 & 9 & .06 & 1 & .01 & 5 & .03 \\
Disgust & 67 & .47 & 134 & .94 & 40 & .28 & 2 & .01 & 14 & .10 & 11 & .08 & 24 & .17 \\
Fear & 129 & .90 & 13 & .09 & 121 & .85 & 4 & .03 & 43 & .30 & 18 & .13 & 66 & .46 \\
Guilt & 55 & .38 & 132 & .92 & 36 & .25 & 0 & .00 & 133 & .93 & 88 & .62 & 11 & .08 \\
Joy & 139 & .97 & 140 & .98 & 4 & .03 & 141 & .99 & 65 & .45 & 41 & .29 & 25 & .17 \\
Sadness & 122 & .85 & 112 & .78 & 88 & .62 & 1 & .01 & 7 & .05 & 2 & .01 & 97 & .68 \\
Shame & 32 & .22 & 111 & .78 & 51 & .36 & 1 & .01 & 106 & .74 & 67 & .47 & 12 & .08 \\
      \cmidrule(r){1-1}\cmidrule(lr){2-3}\cmidrule(rl){4-5}\cmidrule(rl){6-7}\cmidrule(rl){8-9}\cmidrule(rl){10-11}\cmidrule(rl){12-13}\cmidrule(l){14-15}
Total    & 673   &       & 761   &       & 400   &       & 149   &       & 377   &       & 228   &       & 240 & \\
      \toprule
    \end{tabular}
  \caption{Instance counts and ratios across emotions and appraisal annotations.}
  \label{tab:counts}
\end{table}

\subsection{Analysis}
In Table~\ref{tab:counts} are the cooccurrence counts across emotion
and appraisal dimension pairs, as well as the relative counts
normalized by emotion (enISEAR provides 143 descriptions per
emotion). The most frequently annotated class is \certainty, followed
by \attention. Appraisal dimensions are differently distributed across
emotions: \anger and \fear require \attention, \guilt and \shame do
not; \disgust and \anger show the highest association with \certainty,
in opposition to \fear. \emph{Responsibility} and \control play the
biggest role in \guilt and \shame, while \joy, non-surprisingly,
strongly relates to \pleasantness.  \emph{Fear} has a clear link with
\effort and, together with \sadness, it is characterized by the inability
to control the circumstance.

These numbers are particularly interesting in comparison with the
findings of \newcite{Smith1985}, who report the average scores
along the appraisal dimensions (based on a principle component analysis) for each
emotion\footnote{We report the subset of emotions that
  overlap with ours. Also note that their ``Control'' corresponds
  to our ``Circum.''.}. Results are
consistent in most cases. For instance, \joy (or \emotionname{happiness} in
Table~\ref{tab:smithelssworth}) stands out as highly pleasant and
barely related to \effort. Self \responsibility is lowest in \anger,
an emotion that arises when blame is externalized, and mostly present
in \shame and \guilt, which derive from blaming the self
\cite{tracy2006}. These two are also the emotions that annotators
associated with \control more than others. \emotionname{Attention} is
prominent for events that elicited \anger and which were under the
control of others, as suggested by the low \situationalControl. The
highest \situationalControl, on the contrary, appears with
the data points labeled as \fear, also characterized by a strong
feeling of \emotionname{uncertainty} and \effort.
There are also dissimilarities between the two tables, like the level
of attention, reaching the lowest score for \disgust in their study and
not in ours. They also find that situational control is a stronger
indicator for \shame than for \guilt, while \effort is more marked in our
sadness-related events than theirs.

These differences may partly be
data-specific, partly due to the type of metrics shown in the
tables. Most importantly, they can be traced to the annotation setup:
while their subjects recalled and appraised personal events, our
annotators evaluated the descriptions of events that are foreign to
them.

It should be noted that for a reader/annotator it is challenging to
impersonate in the writer: although some events have a shared
understanding (e.g., ``I passed the exam" is most likely appraised as
pleasantness), others are tied to one's personal background, values
and preferences. This may represent a source of disaccord both between
the tables, among the annotators, and with the emotion labels
themselves (e.g., ``I felt ... when my mom offered me curry" has a
\emotionname{pleasant} gold label, while the original author meant it
as a negative emotion, namely \disgust).

\section{Experiments}
\label{sec:experiments}

We now move to our evaluation if automatic methods to recognize emotions can benefit from
being informed about appraisal dimensions explicitly. We first
describe our models and how we address our research questions and then
turn to the results.

\subsection{Model Configuration}
\begin{figure}[t]
  \centering
  \begin{minipage}[b]{0.57\linewidth}
  \centering
  \includegraphics[scale=0.98]{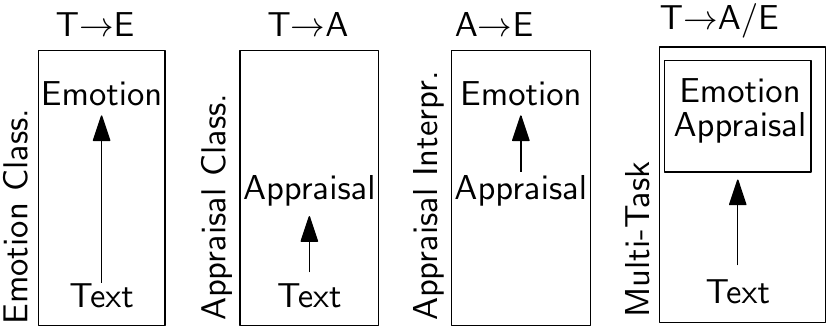}
  \captionof{figure}{Tasks investigated in experiments on appraisal-driven emotion analysis.}
  \label{fig:tasks}
  \end{minipage}
\hfill
\begin{minipage}[b]{0.35\linewidth}
  \centering\small
  \setlength{\tabcolsep}{9pt}
  \renewcommand{\arraystretch}{0.99}
\begin{tabular}{lccc}
  \toprule
  & \multicolumn{3}{c}{\taskTA}\\
  \cmidrule(r){2-4}\cmidrule(r){2-4}
  Appraisal & P & R & \F\\
  \cmidrule(r){1-1}\cmidrule(r){2-4}
  Attention      & 81 &84 &82 \\
  Certainty      & 84 &86 &85 \\
  Effort         & 68 &68 &68 \\
  Pleasantness   & 79 &63 &70 \\
  Responsibility & 74 &68 &71 \\
  Control        & 63 &49 &55 \\
  Circumstance   & 65 &58 &61 \\
  \cmidrule(r){1-1}\cmidrule(l){2-4}
  Macro \avg & 73 &68 &70 \\
  Micro \avg & 77 &74 &75 \\
  \toprule
\end{tabular}
\captionof{table}{Classifier performance on predicting
  appraisal dimensions.}
\label{tab:exp1}
\end{minipage}
\end{figure}

Figure~\ref{fig:tasks} illustrates the four different tasks addressed
by our models.  \textbf{Task \taskTE} is the prediction of
\emph{emotions from text}, namely the standard setting in nearly all
previous work of emotion analysis. We use a convolutional neural
network (CNN) inspired by \newcite{Kim2014}, with pretrained GloVe
(Glove840B) as a 300-dimensional embedding layer
\cite{Pennington2014}\footnote{https://nlp.stanford.edu/projects/glove/}
with convolution filter sizes of 2, 3, and 4 with a ReLu activation
function \cite{Nair2010}, followed by a max pooling layer of length 2
and a dropout of 0.5 followed by another dense layer.

As another model to predict emotions, we use a pipeline based on two
steps, one to detect the appraisal from text, and the second to assign
the appropriate emotion to the appraisal. We refer to \textbf{Task
  \taskTA} as the step of identifying \emph{appraisal from text}. We
use the model configuration of Task \taskTE, except for the sigmoid
activation function and binary cross entropy loss instead of softmax
and cross-entropy loss.
As a second step, \textbf{Task \taskAE} predicts \emph{emotion from
  appraisal}. The features are seven boolean variables corresponding
to the appraisal dimensions. We use a
neural network with two hidden layers with ReLU activation, followed
by a dropout of 0.5.\footnote{We do not perform any further
  hyperparameter search. We also compared all model configurations to
  MaxEnt models to ensure that we do not suffer from overfitting due
  to a large number of parameters, in comparison to the limited
  training data. In all settings, the neural models were superior. We
  therefore limit our explanations to those.}

A disadvantage of the pipeline setting could be that the emotion
prediction needs to handle propagated errors from the first step,
and that the first step cannot benefit from what the second model
learns. Therefore, we compare the pipeline setting (\taskTA, \taskAE)
with a multi-task learning setting (\textbf{\taskTAE}). The model is
similar to Task \taskTE. The convolutional layer is shared by the
tasks of predicting \emph{emotions from text} and predicting
\emph{appraisal from text} and we use two output layers, one for
emotion predictions with softmax activation and one for appraisal
predictions with sigmoid activation.

\begin{table}[t]
\centering\small
\begin{tabular}{lccc ccc ccc ccc ccc ccc}
  \toprule
  &&&&&&&&&&&&&\multicolumn{6}{c}{Oracle Ensembles}\\
  \cmidrule(l){14-19}
  & \multicolumn{3}{c}{\taskTE} & \multicolumn{3}{c}{\taskTA,\taskAE}
  & \multicolumn{3}{c}{\taskAE (Gold)}& \multicolumn{3}{c}{\taskTAE}&
  \multicolumn{3}{c}{T\shortarrow{}A\shortarrow{}E + T\shortarrow{}E}&\multicolumn{3}{c}{T\shortarrow{}A/E + T\shortarrow{}E}\\
  \cmidrule(r){2-4}\cmidrule(rl){5-7}\cmidrule(rl){8-10}\cmidrule(rl){11-13}\cmidrule(lr){14-16}\cmidrule(l){17-19}
  Emotion  & P & R& \F & P & R & \F & P & R & \F & P & R & \F & P & R & \F  & P & R & \F \\
  \cmidrule(r){2-4}\cmidrule(rl){5-7}\cmidrule(rl){8-10}\cmidrule(rl){11-13}\cmidrule(l){14-16}\cmidrule(l){17-19}
  Anger    & 51 &52 &52 & 34 &62 &44 & 55 &71 &62 & 51& 52& 52 & 66&81&73 &  59  & 59 & 59  \\
  Disgust  & 65 &63 &64 & 59 &34 &43 & 53 &48 &51 & 64& 64& 64 & 78&68&73 &  69  & 66 & 67  \\
  Fear     & 69 &71 &70 & 55 &55 &55 & 79 &78 &78 & 70& 68& 69 & 76&77&77 &  73  & 75 & 74  \\
  Guilt    & 47 &42 &44 & 38 &50 &43 & 57 &70 &63 & 45& 42& 44 & 60&63&62 &  58  & 54 & 56  \\
  Joy      & 74 &80 &77 & 77 &69 &72 & 94 &98 &96 & 77& 77& 77 & 79&80&80 &  79  & 85 & 82  \\
  Sadness  & 69 &67 &68 & 58 &40 &47 & 69 &63 &66 & 68& 68& 68 & 74&70&72 &  73  & 71 & 72  \\
  Shame    & 44 &45 &45 & 36 &24 &29 & 56 &35 &43 & 43& 43& 43 & 58&51&54 &  51  & 52 & 52  \\
  \cmidrule(r){2-4}\cmidrule(rl){5-7}\cmidrule(rl){8-10}\cmidrule(rl){11-13}\cmidrule(rl){14-16}\cmidrule(l){17-19}
  Macro \avg&60 &60 &60 & 51 &48 &48 & 66 &66 &65 & 60& 59& 59 &70&70&70 & 66 & 66 & 66 \\
  Micro \avg&   &   &60 &    &   &48 &    &   &66 &   &   & 59 &  &  &70 &    &    & 66 \\
  \toprule
\end{tabular}
\caption{Comparison of the Text-to-Emotion baseline (\taskTE) with the
  performance of first prediction appraisal followed by emotion
  analysis (\taskTA,\taskAE) and the multi-task setting
  (\taskTAE). The oracle consists of a combination of two models and
  is informed which model is more likely to make the correct
  prediction.}
\label{tab:exp2exp3}
\end{table}

\subsection{Results}
\label{sec:results}
We perform each experiment in a repeated 10$\times$10-fold
cross-validation setting and report average results. All partitions
of training and test sets are the same across all experiments.

\textbf{Experiment 1 (Appraisal Prediction, \taskTA)} aims at
understanding how well appraisal dimensions can be predicted from
text. Emotion classification is an established task, and one might
have some intuition on the expected performance with a given data set,
but the prediction of appraisal dimensions has never been performed
before. Hence, we report precision, recall, and \F for each appraisal
component considered in Table~\ref{tab:exp1}. The prediction of
\certainty works best (85\%\F) followed by \attention (82\%\F). The
lowest performance is seen for \control (55\%\F) and \circumstance
(61\%\F). These results are only partially in line with the
inter-annotator agreement scores. We obtain a .75 micro average \F
score.

\newcommand{\mc}[1]{\multicolumn{3}{l}{#1}}
\newcommand{\tbo}[1]{\multirow{2}{*}{\parbox{\linewidth}{#1}}}
\newcommand{\tboo}[1]{\multirow{3}{*}{\parbox{\linewidth}{#1}}}
\newcommand{\sep}{\cmidrule(r){2-2}\cmidrule(r){3-9}\cmidrule{10-10}}

\begin{table}[t]
  \centering\small
  \setlength{\tabcolsep}{2pt}
  \renewcommand{\arraystretch}{0.89}
  \begin{tabularx}{\linewidth}{l|l@{\hskip 10pt}ccccccc@{\hskip 10pt}X}
    \toprule
    \multicolumn{2}{c}{}& \multicolumn{7}{c}{Appraisal} & \\
    \cmidrule(r){3-9}
    \multicolumn{1}{c}{}& Emotion (G/P) & A & Ce & E\phantom{x} & P\phantom{x} & R\phantom{x} & Co & Ci & Text\\
    \sep
    \multirow{7}{*}{\rt{Appr+Emo corr.}}
    &Anger    & 1&1&0&0&0&0&0 & when my neighbour started to throw rubbish in my garden for no reason. \\
    &Disgust  & 0&1&0&0&0&0&0 & to watch someone eat insects on television. \\
    &Fear     & 1&0&1&0&0&0&1 & when our kitten escaped in the late evening and we thought he was lost. \\
    &Guilt    & 0&1&0&0&1&1&0 & when I took something without paying. \\
    & Joy     & 1&1&0&1&1&0&0 & when I found a rare item I had wanted for a long time. \\
    & Sadness & 1&1&1&0&0&0&1 & when my dog died. He was ill for a while. Still miss him.\\
    & Shame   & 0&1&0&0&1&0&0 & when I remember an embarrassing social faux pas from my teenage years. \\
    \midrule
    \multirow{7}{*}{\rt{Emo incorr.}}
    & Anger/Fear   & 1&0&1&0&0&0&0 & when someone drove into my car causing damage and fear to myself -- then drove off before
      exchanging insurance details. \\
    & Disgust/Anger & 1&1&0&0&0&0&0  & when I saw a bird being mistreated when on holiday. \\
    & Fear/Sadness    & 1&1&1&0&0&0&1 & a huge spider just plopped on down on the sofa besides me, staring me out. \\
    & Guilt/Disgust   & 0&1&0&0&0&0&0 & when I watched a
      documentary that showed footage of farms of pigs and chickens
      and as a meat eater I felt awful guilt at how they are treated.
    \\
    & Sadness/Anger & 1&1&0&0&0&0&0 & when I saw a group of homeless people and it was cold outside.\\
    & Shame/Guilt   & 0&1&0&0&1&1&0 & because I did something silly.\\
   \midrule
    \multirow{7}{*}{\rt{Ap+Emo incorr.}}
    & Anger/Shame   & \textbf{0}&1&0&0&\textbf{1}&0&0 & I feel ... because I can't stand when people lie.\\
    & Disgust/Anger & \textbf{1}&1&0&0&0&\textbf{0}&0 & when I saw a medical operation on a TV show.\\
    & Fear/Guilt    & 1&0&\textbf{0}&0&\textbf{1}&0&\textbf{0} & when I was on a flight as I am ... of flying.\\
    & Guilt/Shame   & 0&1&\textbf{1}&0&1&\textbf{1}&0 & when I lost my sister's necklace that I had borrowed. \\
    & Joy/Anger     & 1&1&0&\textbf{0}&0&0&\textbf{0} & when I saw bees coming back to my garden after few years of absence.\\
    & Sadness/Guilt & \textbf{1}&1&0&0&\textbf{1}&0&\textbf{0} & when I watched some of the sad cases of children in need.\\
    & Shame/Guilt   & 0&1&0&0&1&\textbf{1}&0 & when I forgot a hairdressers appointment.\\
    \toprule
  \end{tabularx}
  \caption{Examples for the prediction of the pipeline setting
    (\taskTA, \taskAE) . A: Attention, Ce: Certainty, E: Effort, P:
    Pleasantness, R: Responsibility, Co: Control, Ci:
    Circumstance. First emotion mention is gold, second is prediction.
    Appraisal shown is prediction with errors shown in bold.
  }
  \label{tab:analysis_task2_3}
\end{table}

\textbf{Experiment 2 (Appraisal Interpretation, \taskAE)} aims at
understanding how well emotions can be predicted from appraisals.  We
compare the baseline text-to-emotion setting (\taskTE) to the pipeline
setting that first predicts the appraisal and then, from those, the
emotion.  In the pipeline setting we train the second step (\taskAE)
on the gold appraisal annotations, not on the predictions.\footnote{We
  also tested if training on the prediction leads to better results,
  but it constantly underperformed.} We compare this setting to the
performance of the appraisal-to-emotion model (\taskAE), when applied
on gold appraisal annotations. This serves as an upper bound which can
be reached with the best-performing appraisal prediction model.

We first turn to the results of the model which predicts the emotion
based on annotated (gold) appraisal dimensions (\taskAE (gold)). Here,
we observe a clear improvement in contrast to the emotion
classification which has access to the text (\taskTE). \emph{Anger}
increases from .52 to .62, \disgust decreases from .64 to .51, \fear
increases from .70 to .78, \guilt from .44 to .63, \joy from .77 to
.96, \sadness decreases from .68 to .66 and \shame decreased from .45
to .43.
On micro average, the performance increases from .60 to .66\% \F. These
results are an upper-bound for the performance that can be achieved
with the pipeline model, under the assumption of having access to
perfect appraisal predictions.

When moving to the real-world setting of first predicting the
appraisal dimensions and then, based on those, predicting the emotion,
the performance scores drop from 66 to 48\% \F. This is an indicator
that the performance of our current appraisal prediction, though
comparably reasonable with 75\% \F, is not yet sufficient to support
emotion predictions, at least partially. The clear improvement in
emotion prediction based on perfect appraisal annotations and the
performance drop in the real-world setting for emotion prediction
suggest that annotating more data with appraisal dimensions is
necessary to further develop our approach.

Finally, \textbf{Experiment 3 (Multi-Task Learning, \taskTAE and
  Oracle Ensembles)} evaluates if the model which learns appraisals and
emotions jointly performs better than the pipeline model (both being
models applicable in a real-world scenario). We show these results
also in Table~\ref{tab:exp2exp3} and see that the multi-task learning
model cannot improve over the text-only setting.

A remaining question is on the complementarity of the pipeline and the
multitask model to the \taskTE model. To look into this, we define two
oracle ensembles (\taskTA, \taskAE and \taskTE as well as \taskTAE and
\taskTE), in which an oracle predicts which of the two approaches will
obtain the correct result. In this experiment, we therefore accept a
prediction as true positive if one of the two parts of the ensemble is
correct. These results are shown in Table~\ref{tab:exp2exp3} in the
columns ``Oracle Ensemble''.  We see a clear improvement over the
isolated text-based prediction for both oracle ensembles, while the
pipeline model shows a higher contribution in addition to the \taskTE
model (70\% \F in contrast to 66\% \F).

\subsection{Discussion and Analysis}
We have seen in the experiments and results that the approach of
predicting emotions based on appraisal shows a clear potential for a
better performance. Though we have not been able to reach a substantial
improvement in a real-world setting, in which appraisal dimensions are
first predicted as a basis for the emotion prediction or in the
multi-task setting, we observe that text- and appraisal-based models
behave differently.
Table~\ref{tab:analysis_task2_3} shows examples for the prediction in
the real-world setting (\taskTA, \taskAE). In the top block, one
example is shown for each emotion in which both appraisal and emotion
are correctly predicted. This does not only include cases in which
clear emotion indicators exist.
The second block reports instances in which the appraisal is correct, but
the emotion prediction is not. Here, the first sentence (``when someone
drove into my car\ldots'') is an example in which a flip of \certainty
would have changed the emotion. Similarly, the change of \attention in
``when I saw a group\ldots'' would have lead to the correct emotion
prediction. These are therefore untypical cases of appraisal
assignment.
The last block shows examples where the wrong appraisal
prediction leads to wrong emotion assignment.

It is further interesting to look into those cases which are wrongly
predicted from text, but correctly predicted based on the gold
appraisal annotations. We show examples for such cases in
Table~\ref{tab:analysis_gold}. Several of these cases are examples in
which a word seems to indicate a particular emotion, which is actually
not relevant to infer the emotion in the first place (e.g.,
``animal'', ``vomiting'', ``kids'', ``high school''). Often, \shame is
wrongly predicted when the event is about the self. This is
particularly problematic if the actual word pointing to an emotion
appears to be non-typical (e.g., ``crossword'', ``anaesthetic'').

\begin{table}
  \centering\small
  \setlength{\tabcolsep}{6pt}
  \renewcommand{\arraystretch}{0.9}
  \begin{tabular}{lll p{100mm}}
      \toprule
      Gold Emotion & \taskAE & \taskTE & Text\\
      \cmidrule(l){1-1}\cmidrule(l){2-2}\cmidrule(l){3-3}\cmidrule(l){4-4}
      Anger & Anger & Fear      & because I was overlooked at work. \\
      Anger & Anger & Disgust   & when I saw someone mistreating an animal. \\
      Anger & Anger & Fear      & when someone overtook my car on a blind bend and nearly caused an accident. \\
      Disgust & Disgust & Shame & because I ate a sausage that was horrible. \\
      Disgust & Disgust & Fear  & when I was on a ferry in a storm and lots of people were vomiting. \\
      Disgust & Disgust & Shame & because the milk I put in my coffee had lumps in it. \\
      Fear & Fear & Shame       & because I had to have a general anaesthetic for an operation. \\
      Fear & Fear & Sadness     & when my 2 year old broke her leg, and we felt helpless to assist her. \\
      Fear & Fear & Anger       & because we were driving fast in the rain in order to get somewhere before it shut, and the driver was going over the speed limit. \\
      Guilt & Guilt & Shame     & when I took something without paying. \\
      Guilt & Guilt & Joy       & for denying to offer my kids what they demanded of me. \\
      Guilt & Guilt & Anger     & when I had not done a job for a friend that I had promised to do. \\
      Joy & Joy & Sadness       & when witnessing the joy on my children's face on Christmas morning. \\
      Joy & Joy & Shame         & when I managed to complete a cryptic crossword. \\
      Joy & Joy & Disgust       & when I found a twenty pound note on the ground outside. \\
      Sadness & Sadness & Fear  & when it was raining this morning as I been planning to go on a camping trip. \\
      Sadness & Sadness & Joy   & I feel ... when I see the Christmas decorations come down, and know they won't be up again for another year. \\
      Sadness & Sadness & Shame & when my friend's eye was watering after an injection into it and I could do nothing to help. \\
      Shame & Shame & Joy       & when I failed my ninth year at high school. \\
      Shame & Shame & Guilt     & when I had too much to drink in a pub, fell over and had to go to hospital. \\
      Shame & Shame & Anger     & when my mom caught me lying. \\
      \toprule
  \end{tabular}
  \caption{Examples in which the appraisal model (on gold
    appraisal annotation) predicts the correct emotion and the
    baseline system does not.}
  \label{tab:analysis_gold}
\end{table}

\section{Conclusions and Future Work}
\label{sec:conclusion}
We investigated the hypothesis that informing an emotion
classification model about the cognitive appraisal regarding a
situation is beneficial for the model performance. We were able to
show that emotion classification performs better than text-based
classification, under the assumption that perfect appraisal
predictions are possible and shows complementary correct
predictions. Yet, neither in a multi-task learning nor a
pipeline, in which the appraisal was predicted as a basis, could we
show an improvement in emotion classification. This provides evidence that, though
our appraisal predictor is of reasonable performance, the model
suffers from error propagation. This is still an encouraging result,
suggesting that future work should further investigate the combination
of appraisal information with emotion prediction, particularly in the
light of our oracle ensemble that indicated a clear improvement.

This first study on the topic raises a couple of research questions:
Would there be other neural architectures which are better suited for
including the appraisal information? Will more annotated data improve
the prediction quality sufficiently? Finally, it should be analyzed if
giving the annotators access to the emotion label when making the
appraisal annotation could have changed the results.

\section*{Acknowledgements}
This work was supported by Leibniz WissenschaftsCampus T\"ubingen
``Cognitive Interfaces'' and Deutsche Forschungsgemeinschaft (project
SEAT, KL 2869/1). We thank Laura Ana Maria Oberl\"ander for
inspiration and fruitful discussions and Valentino Sabbatino for his
annotation work. Further, we thank the three anonymous reviewers for
their constructive and helpful criticism.

\bibliographystyle{acl}
\bibliography{lit}

\appendix

\section{Appraisal Prediction Performance of the Multitask Model}
\begin{wraptable}[15]{R}{0.3\textwidth}
  \centering\vspace{-2\baselineskip}
  \begin{tabular}{lccc}
    \toprule
    & \multicolumn{3}{c}{\taskTAE}\\
    \cmidrule(r){2-4}\cmidrule(r){2-4}
    Appraisal & P & R & \F\\
    \cmidrule(r){1-1}\cmidrule(r){2-4}
    Attention      & 82 &78 &80 \\
    Certainty      & 85 &76 &80 \\
    Effort         & 65 &62 &64 \\
    Pleasantness   & 76 &59 &67 \\
    Responsibility & 71 &69 &70 \\
    Control        & 60 &42 &49 \\
    Circumstance   & 67 &48 &56 \\
    \cmidrule(r){1-1}\cmidrule(l){2-4}
    Macro \avg & 72 &62 &67 \\
    Micro \avg & 76 &68 &71 \\
    \toprule
  \end{tabular}
  \captionof{table}{Multi-task learning model performance on predicting
    appraisal dimensions.}
  \label{tab:exp1_mtl}
\end{wraptable}
We focused in the discussion of the results in
Section~\ref{sec:results} on the goal to improve the performance of
the emotion prediction with appraisal information. Though we evaluated
the appraisal prediction model as an intermediate step in the pipeline
to emotion prediction, we did only evaluate the emotion performance in
the multitask learning setting due to our focus on emotion
categorization.

However, it would be meaningful to also experiment with a
T$\rightarrow$E,E$\rightarrow$A model when the goal would be to
predict the appraisal dimensions with or without knowledge of an
emotion category. We leave such evaluation for future work. However,
our multitask model \taskTAE also produces appraisal predictions in
the context of emotion predictions. We therefore provide the appraisal
evaluation of the \taskTAE model in Table~\ref{tab:exp1_mtl}. We see
that the appraisal prediction does not show any improvement over the
\taskTA model with the respective results in Table~\ref{tab:exp1}.

\section{Examples from the Multitask Model}
We show examples from the \taskTAE model in
Table~\ref{tab:analysis_task2_3_mtl}, similarly to the results from
\taskTA,\taskAE examples in Table~\ref{tab:analysis_task2_3}. Note
that, given the bidirectional interdependencies between emotion and
appraisal taking place in the multitask learning, in contrast to the
unidirectional information flow from appraisal to emotion in the
pipeline, these results are more challenging to interpret.\\

\begingroup
  \centering\small
  \setlength{\tabcolsep}{2pt}
  \renewcommand{\arraystretch}{0.95}
  \begin{tabularx}{16cm}{l|l@{\hskip 10pt}ccccccc@{\hskip 10pt}X}
    \toprule
    \multicolumn{2}{c}{}& \multicolumn{7}{c}{Appraisal} & \\
    \cmidrule(r){3-9}
    \multicolumn{1}{c}{}& Emotion (G/P) & A & Ce & E\phantom{x} & P\phantom{x} & R\phantom{x} & Co & Ci & Text\\
    \sep
    \multirow{7}{*}{\rt{Appr+Emo corr.}}
    & Anger   & 1&1&0&0&0&0&0 & when I heard someone gossip about a friend of mine. \\
    & Disgust & 0&1&0&0&0&0&1 & because the drain smelled. \\
    & Fear    & 1&0&1&0&0&0&0 & when I thought my son could be in trouble when mountain climbing. \\
    & Guilt   & 0&1&0&0&1&1&0 & because I drank way too much, knowing that I had things to do the next day. \\
    & Joy     & 1&1&0&1&0&0&0 & because my son gave me a big hug when I got home from work. \\
    & Sadness & 1&1&0&0&0&0&0 & when I read about orangutans and palm oil \\
    & Shame   & 0&1&0&0&1&1&0 & because I walked past a beggar and didn't bother to stop and give her a pound. \\
    \midrule
    \multirow{7}{*}{\rt{Emo incorr.}}
    & Anger/Shame   & 1&0&0&0&0&0&0 & because the parents dropping their kids off to
        school were driving and parking dangerously.\\
    & Anger/Disgust   & 1&0&1&0&0&0&0 & someone stabbed a man on a train. \\
    & Disgust/Sadness & 1&1&0&0&0&0&0 & when I read that hunters had killed one of the world famous lions. \\
    & Disgust/Anger   & 1&1&0&0&0&0&0 & when I watched videos of how some people treat animals. \\
    & Guilt/Shame     & 0&1&0&0&1&1&0 & because I cheated on a previous girlfriend. \\
    & Sadness/Anger   & 1&1&0&0&0&0&0 & when listening to the national news. \\
    & Shame/Guilt     & 0&1&0&0&1&1&0 & when my daughter was rude to my wife. \\
   \midrule
    \multirow{7}{*}{\rt{Ap+Emo incorr.}}
    & Anger/Disgust & \textbf{1}&1&\textbf{0}&0&0&0&0 & when a drunk man insulted my intelligence. \\
    & Disgust/Fear  & 1&\textbf{1}&\textbf{0}&0&\textbf{1}&0&\textbf{0} & towards my friend because he stole my wallet. \\
    & Fear/Shame    & \textbf{0}&\textbf{1}&\textbf{0}&0&\textbf{1}&0&0 &  because a fight broke out in a bar where I was drinking \\
    & Guilt/Disgust & \textbf{0}&1&\textbf{0}&0&\textbf{0}&0&0 & because I accidentally killed my pet fish by decorating his tank with untreated sand.\\
    & Joy/Sadness   & 1&1&0&\textbf{0}&\textbf{0}&\textbf{0}&0 & when I rescued a dog at the local shelter and he became my best friend. \\
    & Sadness/Joy & \textbf{1}&1&\textbf{0}&0&0&0&\textbf{1} & when I was on holiday and it was the last day and had to come home. \\
    & Shame/Guilt    & 0&1&0&0&1&\textbf{0}&1 & when I ate my whole Easter egg in one go.\\
    \toprule
  \end{tabularx}
  \captionof{table}{\label{tab:analysis_task2_3_mtl} Examples for the prediction of the multi-task setting
    (\taskTAE) . A: Attention, Ce: Certainty, E: Effort, P:
    Pleasantness, R: Responsibility, Co: Control, Ci:
    Circumstance. First emotion mention is gold, second is prediction.
    Appraisal shown is prediction with errors shown in bold.
  }

\endgroup

\end{document}